%% file: main.tex
\documentclass[letterpaper, 10 pt, conference]{ieeeconf}  %

\IEEEoverridecommandlockouts                              %

\usepackage{ral}
\usepackage{url}
\usepackage{fontawesome5}

\title{\LARGE \bf Learning Acrobatic Flight from Preferences}

\author{Colin Merk$^{*,1}$, Ismail Geles$^{*,1}$, Jiaxu Xing$^{1}$, Angel Romero$^{1}$, Giorgia Ramponi$^{2}$, and Davide Scaramuzza$^{1}$%
\thanks{*These authors contributed equally to this work.}%
\thanks{$^{1}$These authors are with the Robotics and Perception Group, University of Zurich, Switzerland (\url{https://rpg.ifi.uzh.ch}). \linebreak Contact: \texttt{geles@ifi.uzh.ch}.}%
\thanks{$^{2}$Giorgia Ramponi is with the Autonomous Learning and Predictive Intelligence Lab, University of Zurich, Switzerland.}%
\thanks{Supported by the European Union’s Horizon Europe programme (grant No.~101120732, AUTOASSESS) and the European Research Council (ERC) (grant No.~864042, AGILEFLIGHT).}%
}

\usepackage{eso-pic}
\usepackage{xcolor}
\definecolor{somegray}{rgb}{0.5, 0.5, 0.5}
\newcommand\PaperNotice{%
\AddToShipoutPictureFG*{%
    \AtPageUpperLeft{%
        \put(0,-40){ %
            \parbox{\paperwidth}{%
                \centering
                \color{somegray}\large
                This paper has been accepted for publication at the
                IEEE/RSJ International \\ Conference on Intelligent Robots and Systems (IROS), Pittsburgh, 2026. ©IEEE
            }
        }
    }
}
}

\begin{document}
\PaperNotice

\maketitle
\thispagestyle{empty}
\pagestyle{empty}

\begin{abstract}
    Preference-based reinforcement learning (PbRL) enables agents to learn control policies without requiring manually designed reward functions, making it well-suited for tasks where objectives are difficult to formalize or inherently subjective. Acrobatic flight poses a particularly challenging problem due to its complex dynamics, rapid movements, and the importance of precise execution. 
    However, manually designed reward functions for such tasks often fail to capture the qualities that matter: we find that hand-crafted rewards agree with human judgment only 60.7\% of the time, underscoring the need for preference-driven approaches. In this work, we propose Reward Ensemble under Confidence (REC), a probabilistic reward learning framework for PbRL that explicitly models per-timestep reward uncertainty through an ensemble of distributional reward models. By propagating uncertainty into the preference loss and leveraging disagreement for exploration, REC achieves 88.4\% of shaped reward performance on acrobatic quadrotor control, compared to 55.2\% with standard Preference PPO. We train policies in simulation and successfully transfer them zero-shot to the real world, demonstrating complex acrobatic maneuvers learned purely from preference feedback. We further validate REC on a continuous control benchmark, confirming its applicability beyond the domain of aerial robotics.
\end{abstract}

\input{chapters/introduction}

\input{chapters/relatedwork}

\input{chapters/method}

\input{chapters/results}

\input{chapters/conclusion}

\bibliographystyle{ieeeTran}
\bibliography{bibliography}

\input{chapters/appendix}

\end{document}

%% file: chapters/introduction.tex
\vspace{-0.2cm}

\section*{Supplementary Materials}

\noindent
\faVideo\hspace{0.6em}\text{Video:}  
\url{https://youtu.be/JiMW_ligKvQ}

\section{Introduction}
\label{sec:introduction}

Autonomous drones capable of executing dynamic and agile maneuvers have become a benchmark for progress in robotics, with applications ranging from high-speed navigation~\cite{kaufmann_champion-level_2023} and search-and-rescue to inspection~\cite{xing2023autonomous}. Reinforcement learning (RL) has shown promise in enabling such high-performance control policies, even surpassing expert human pilots in competitive environments~\cite{kaufmann_champion-level_2023}. However, a persistent challenge in applying RL to real-world tasks is the design of effective reward functions.

Reward design is not only time-consuming and task-specific, but it also introduces a fundamental bottleneck: many tasks---particularly those involving aesthetics, subjective quality, or high-level intent---are difficult or even impossible to express with a well-defined reward function. This limitation is especially critical in acrobatic flight maneuvers, where desirable behavior may depend on human preferences over trajectory smoothness, timing, or style.

\input{figures/tex/eyecatcher_new}

Acrobatic flight compounds this difficulty due to its highly nonlinear dynamics, rapid state transitions, and narrow margins for error. Small differences in execution can distinguish a smooth, visually compelling maneuver from a jerky or failed one, yet formalizing such distinctions in a reward function requires extensive domain expertise and iterative tuning. Even carefully engineered rewards may fail to reflect what observers actually prefer: we show that hand-crafted rewards agree with human judgment only $60.7\%$ of the time.

To address this, preference-based reinforcement learning (PbRL) has emerged as a compelling alternative. Rather than specifying explicit rewards, PbRL infers a reward function from comparisons between trajectory segments~\cite{christiano2023deepreinforcementlearninghuman}. 
This approach enables policy learning for tasks that are hard to define but easy to evaluate by preference. Each acrobatic skill would typically require its own carefully designed reward function in traditional RL, whereas PbRL offers a more general framework that can adapt across different maneuvers without task-specific reward engineering. 
Unlike imitation learning, which requires high-quality demonstrations, PbRL only requires comparative judgments between behaviors, decoupling the quality of the learned policy from the skill level of the feedback source.

Despite the promise of PbRL, its application in real-world robotics has been limited. Most prior work has focused on simulated benchmarks, and applications to physical systems, particularly agile drones, remain scarce. Existing methods also tend to overlook the uncertainty inherent in preferences, which can lead to instability or suboptimal learning when feedback is noisy or sparse. Without accounting for this uncertainty, reward models can overfit to ambiguous preference labels, leading to brittle policies that may fail to deploy.

In this work, we propose \textit{Reward Ensemble under Confidence} (REC), a probabilistic preference-based reinforcement learning framework for acrobatic drone flight. REC introduces an ensemble of distributional reward models that explicitly capture per-timestep reward uncertainty and propagate it through the preference loss via a Gaussian cumulative distribution function.

The key insight is that preference feedback is inherently probabilistic: when two trajectories are similar in quality, the preferred choice is noisy. Modeling each reward as a distribution lets REC express confidence in its predictions, and the resulting ensemble disagreement drives exploration toward reward regions where the model is least certain, yielding more stable training under limited preference supervision.

We validate our approach through simulation and real-world experiments on acrobatic quadrotor control, as well as on a standard continuous control benchmark. Our contributions are as follows:
\begin{enumerate}
    \item We propose REC, a probabilistic reward learning framework that models per-timestep reward uncertainty within an ensemble of reward models, replacing the standard Bradley-Terry softmax with a distributional preference model.
    \item We demonstrate that REC achieves $88.4\%$ of shaped reward performance on acrobatic quadrotor control, compared to $55.2\%$ with standard Preference PPO, representing a substantial improvement in preference-based learning for agile flight.
    \item We successfully transfer learned policies zero-shot to a real 220\,g quadrotor, executing acrobatic maneuvers including continuous powerloops and a novel vertical Figure-8 learned purely from preference feedback.
    \item We show that hand-crafted reward functions agree with human judgment only $60.7\%$ of the time, highlighting the limitations of manual reward engineering for tasks with subjective objectives.
\end{enumerate}

%% file: figures/tex/eyecatcher_new.tex
\begin{figure}
    \centering
    \begin{tikzpicture}[
    scale=0.99,
    node distance=1.2cm and 2cm,
    every node/.style={font=\small},
    box/.style={draw, rectangle, align=center},
    arrow/.style={-{Latex}, thick}, 
    trapezium box/.style={
        draw, trapezium, trapezium angle=70,
        minimum width=2.5cm, minimum height=1cm, rotate=-90, align=center, fill=Emerald!10
    }]
    \node[box, rounded corners, fill=Lavender!10, color=Lavender!10, minimum height=6.9cm, minimum width=8.8cm] (box1) at (2.8, -2) {};

    \node[box, minimum width=2.8cm, minimum height=1cm, rounded corners, text width=1.7cm, align=left, color=Lavender!10, fill=Lavender!10] (T1) {\includegraphics[width=2.5cm]{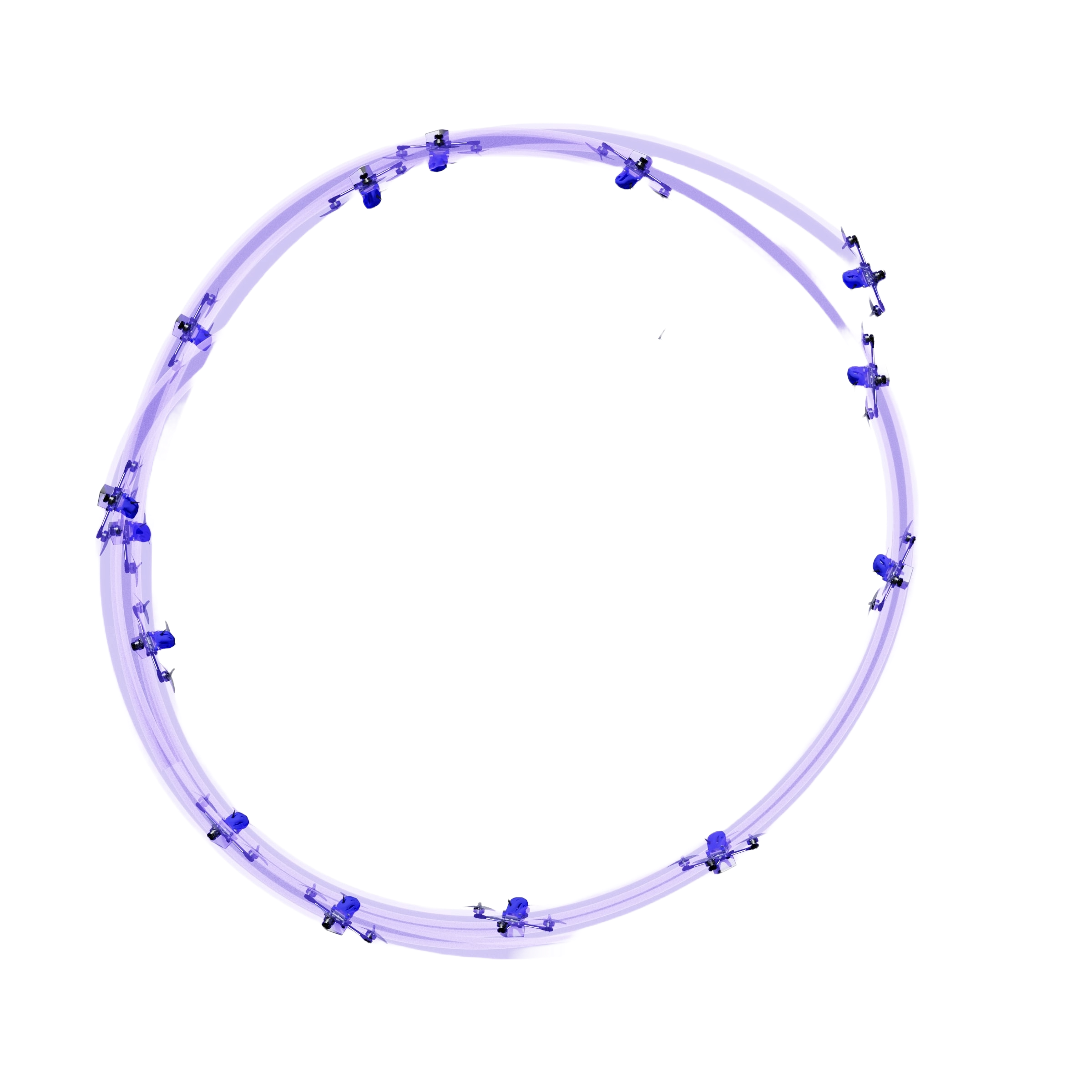}};
    \node[box, minimum width=2.8cm, minimum height=1cm, rounded corners, text width=1.8cm, align=left, color=Lavender!10, fill=Lavender!10, below=-0.2cm of T1] (T2) {\includegraphics[width=2.5cm]{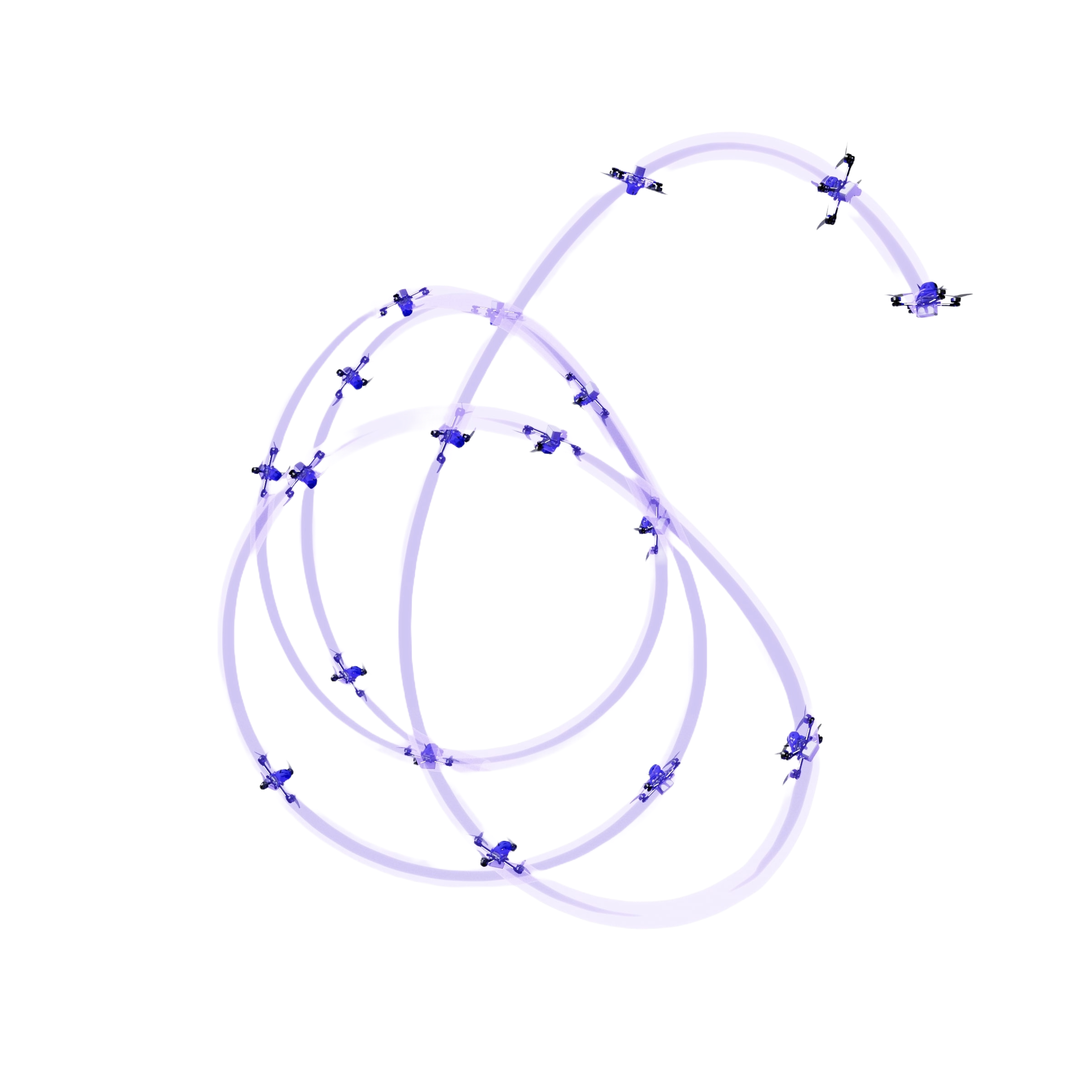}};

    \node[box, minimum width=1.5cm, minimum height=0.8cm, color=Lavender!10,fill=Lavender!10, right=of T1, yshift=-1.1cm, xshift=-2cm] (judge) {\includegraphics[width=2.5cm]{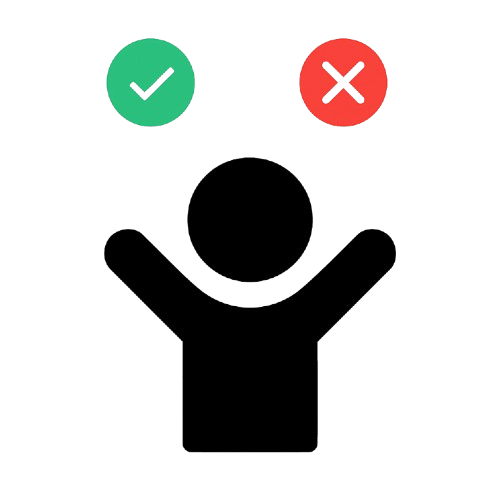}};

    \node[trapezium box, right=1.3cm of judge, xshift=-0.8cm, yshift=0.2cm] (trap1) {};
    \node[trapezium box, right=0.1cm of trap1, xshift=-1.82cm] (trap2) {};
    \node[trapezium box, right=0.1cm of trap2, xshift=-1.82cm] (prefmodel) {};
    \node[box, color=Emerald!10,fill=Emerald!10,  right=0.4cm of trap2, xshift=-1cm, yshift=0.9cm] (gaussian) {\includegraphics[width=1.1cm]{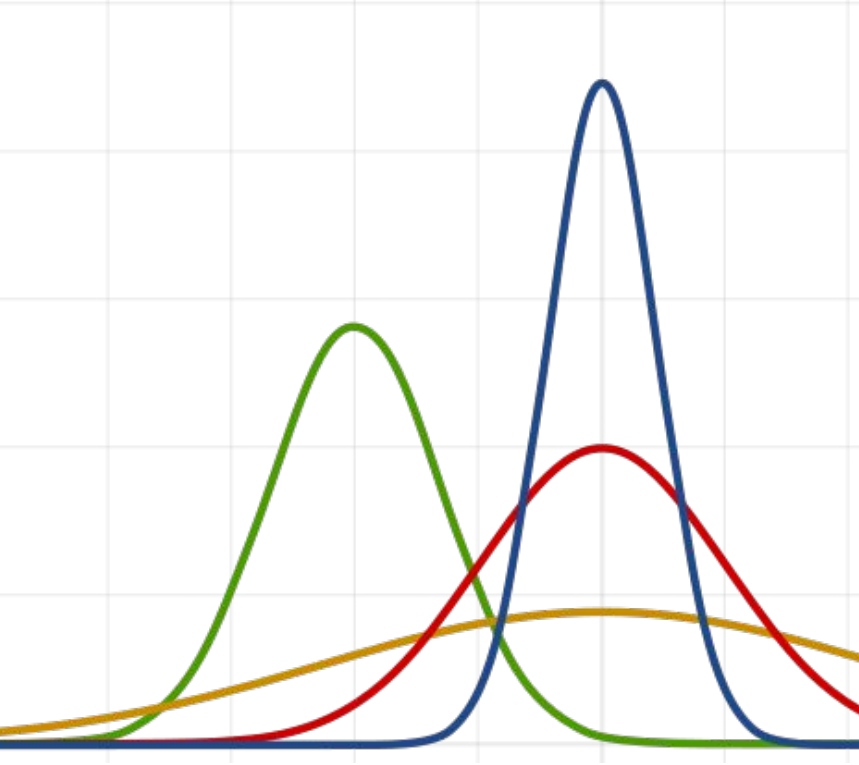}};

    \node[box, rounded corners, minimum width=2.4cm, minimum height=1.2cm, align=center, draw=none, fill=NavyBlue!10, below=2.3cm of prefmodel, xshift=-1.5cm] (actorcritic) {RL Training};

    \node[box, rounded corners, fill=yellow!10, draw=none, minimum width=1.6cm, minimum height=1.2cm, left=1.2cm of actorcritic] (policy) {$\pi_\theta(a|s)$};

    \node(text_sim) at (5.4, 1) {a) Training in \textit{\textcolor{OrangeRed}{Simulation}}};
    \node(text_sim) at (-1.1, 0.0) { $\tau_1$};
    \node(text_sim) at (-1.1, -2.7) { $\tau_2$};

    \node(text_sim) at (2.1, 0.4) { \textcolor{ForestGreen}{$\tau_1$}};
    \node(text_sim) at (3.25, 0.4) { \textcolor{OrangeRed}{$\tau_2$}};
    \node(text_sim) at (2.675, 0.4) { $>$};
    \node[align=center, color=black!75](text_sim) at (2.725, -2.5) {\textit{Better Powerloop?}};
    \node[align=center](rw_model) at (5.8, -3.4) {\textit{Reward} \\Model $r$};

    \draw[arrow, color=black!40] (rw_model.south) -- ++(0, -1 + 0.3) -- ++(-0.8, 0);
    \draw[arrow, color=black!40] (3.6, -2.0+ 0.3) -- ++(1, 0);
    \draw[arrow, color=black!40] (2.2, -4.8+ 0.3) -- ++(-0.8, 0);

    \node[box, rounded corners, fill=NavyBlue!5, color=NavyBlue!5, minimum height=3.5cm, minimum width=8.8cm] (box1) at (2.8, -7.5) {};
     
    \node(text_sim) at (5.1, -6.2) {b) Deployment in \textit{\textcolor{ethblue}{Real World}}};
        
    \node[box, rounded corners, fill=yellow!10, draw=none, minimum width=1.6cm, minimum height=1.2cm, below=2.1cm of policy] (policy1) {$\pi_\theta(a|s)$};
    \node[box, minimum width=1.5cm, minimum height=0.8cm, color=gray!10,fill=gray!10, right=of policy1,  xshift=0.2cm] (judge) {\includegraphics[width=3cm]{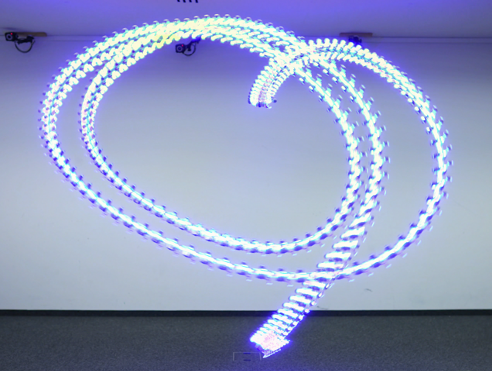}};

    \draw[arrow, color=Orange!40] (policy.south) -- (policy1.north);
    \draw[arrow, color=ethblue!60] (-1.4, -8.2 + 0.3) -- ++(0.8, 0);
    \draw[arrow, color=ethblue!60] (1.7, -8.2 + 0.3) -- ++(1.4, 0);
    \node (obs1) at (-1., -9.1 + 0.7) {{\scriptsize$\bm{p}, \tilde{\bm{R}}, \bm{v},$}};
    \node (obs1) at (-1., -9.5+ 0.7) {\scriptsize$\bm{\omega}, a_\text{prev}$};
    \node (obs1) at (2.4, -9.1+ 0.7) {\scriptsize$c , \bm{\omega}_B^\intercal$};
    \node[align=center, color=Orange!80](sim2real) at (1.3, -7.1+ 0.7) {\textit{Sim2Real}\\ Transfer};
    
    \end{tikzpicture}
    \vspace*{-0.6cm}
    \caption{Overview of the proposed approach. (a) In simulation, trajectory pairs $(\tau_1, \tau_2)$ are presented to an annotator (human or synthetic) to collect preference labels. These labels train an ensemble of reward models with uncertainty estimation, which provides the reward signal for policy optimization via reinforcement learning. (b) The resulting policy is transferred zero-shot to a real quadrotor to execute acrobatic maneuvers.}
    \label{fig:eyecatcher}
    \vspace*{-0.6cm}
\end{figure}

%% file: chapters/relatedwork.tex
\section{Related Works}
\label{sec:relatedworks}

\textbf{Preference-Based Reinforcement Learning.} Preference-based reinforcement learning (PbRL) enables agents to learn policies directly from comparisons between trajectory segments, typically provided by human evaluators, bypassing the need for manually designed reward functions. This approach was introduced by \cite{christiano2023deepreinforcementlearninghuman} and has since inspired numerous advances. Furthermore, limitations in time and sample efficiency have been mitigated through approaches that actively generate pairwise comparisons guided by information gain~\cite{biyik2020askingeasyquestions}, by selecting batches of comparison pairs~\cite{biyik2024batch}, and active volume removal \cite{sadigh2017ActivePL}. Recent efforts include enhancing exploration efficiency through unsupervised pretraining \cite{lee_pebble_2021}, incorporating transformer-based reward models to capture temporal and non-Markovian aspects of human feedback \cite{kim2023preferencetransformermodelinghuman}, and exploring implicit reward learning via Q-functions without explicitly modeled rewards \cite{hejna2023inversepreferencelearningpreferencebased}. Similarly, \cite{liang2022rewarduncertaintyexplorationpreferencebased} leverage reward uncertainty from ensemble disagreement to drive exploration in preference-based learning.

\textbf{Real-World Deployments of PbRL.} Despite its success in simulation, PbRL has seen limited application in real-world robotics, though recent studies show growing interest. One line of research improves sample efficiency by leveraging pre-trained preference models from prior tasks, enabling adaptation to novel tasks on robotic manipulators \cite{hejna2022fewshot}, or using priors obtained from pre-trained LLMs for higher quality candidates in PbRL to achieve expressive behaviors on a quadruped~\cite{clark2024lgpl}. Another approach combines demonstrations with preference queries to address the inefficiencies of standard PbRL and the limitations of inverse reinforcement learning \cite{sadigh2019learningrewardfunctions}.

\textbf{Reinforcement Learning in Quadrotor Control.} Reinforcement learning has significantly advanced quadrotor control capabilities, with remarkable successes in drone racing \cite{kaufmann_champion-level_2023, Song23Reaching}, robust maneuvering under disturbances \cite{chenhuan2021robustquadrotorrl}, and agile flight tasks \cite{lupashin_simple_2010, xing2024multi}. 
Recent works have further demonstrated vision-based RL without using any state information~\cite{geles2024demonstrating, xing2024bootstrapping}. However, most of these methods rely on manually engineered reward functions. To the best of our knowledge, PbRL has not been applied to quadrotor acrobatics. Our work addresses this gap, demonstrating both simulation efficacy and real-world feasibility using preference feedback from both synthetic and human sources.

%% file: chapters/method.tex
\section{Methodology}
\label{sec:methodology}

\subsection{Preliminaries}

We consider a standard RL setting in which an agent takes an action $a_t$ after receiving an observation $o_t$ and obtains a reward $r_t = r(o_t, a_t)$. In traditional RL this reward is hand-crafted, whereas PbRL replaces it with a reward model trained from preference labels over pairs of trajectories. In online PbRL these labels are collected by repeatedly presenting trajectory pairs to an annotator who selects the preferred one, alternating with policy optimization to iteratively improve the agent's behavior.

\textbf{Trajectory.} We denote a trajectory $\tau$ as a sequence of observation–action pairs $(o_i, a_i)$ of length $N$:
\begin{equation}
    \tau = \big((o_1, a_1), (o_2, a_2), \dots, (o_N, a_N)\big),
\end{equation}
where $o_i$ is the observation at timestep $i$ and $a_i$ the corresponding action. The total reward of a trajectory $r(\tau)$ is the sum of the rewards at each timestep:
\begin{equation}
    r(\tau) = \textstyle\sum_{i=1}^N r_i.
\end{equation}

The reinforcement learning objective can be formalized as maximizing the expected sum of discounted rewards:
\begin{equation}
    \max_\pi \; \mathbb{E}_{\tau \sim \pi} \left[ r(\tau) \right]
    = \max_\pi \; \mathbb{E}_{\tau \sim \pi} \left[ \textstyle\sum_{i=1}^N \gamma^{i} r_i \right],
\end{equation}
where $\pi$ denotes the policy, $\tau \sim \pi$ represents trajectories sampled from the policy, and $\gamma \in [0, 1]$ is the discount factor.

\subsection{Quadrotor System Model}
We consider the problem of learning acrobatic flight policies for a quadrotor platform. We use Flightmare~\cite{song2020flightmare} and Agilicious~\cite{foehn2022agilicious} for training in simulation and deploying in the real world. The quadrotor dynamics are modeled as
\begin{align*}
\dot{\bm{x}} =
\begin{bmatrix}
\dot{\bm{p}}_{\wfr\bfr} \\  
\dot{\bm{q}}_{\wfr\bfr} \\
\dot{\bm{v}}_{\wfr} \\
\dot{\boldsymbol\omega}_\bfr \\
\dot{\boldsymbol\Omega}
\end{bmatrix} = 
\begin{bmatrix}
\bm{v}_\wfr \\  
\bm{q}_{\wfr\bfr} \cdot \begin{bmatrix}
0 \\ \bm{\omega}_\bfr/2\end{bmatrix} \\
\frac{1}{m} \Big(\bm{q}_{\wfr\bfr} \odot (\bm{f}_\text{prop} + \bm{f}_\text{aero})\Big)+\bm{g}_\wfr  \\
\bm{J}^{-1}\big( \boldsymbol{\tau}_\text{prop} + \boldsymbol{\tau}_\text{aero}  - \boldsymbol{\omega}_\bfr \times \bm J \boldsymbol{\omega}_\bfr\big) \\
\frac{1}{k_\text{mot}} \big(\boldsymbol\Omega_\text{ss} - \boldsymbol\Omega \big)
\end{bmatrix} \; ,
\end{align*}
where $\odot$ denotes quaternion-based rotation, $\bm{p}_{\wfr\bfr}$, $\bm{q}_{\wfr\bfr}$, $\bm{v}_{\wfr}$, and $\boldsymbol\omega_\bfr$ are the vehicle's position, orientation, velocity, and angular rates, $\bm{J}$ is the inertia tensor, $k_\text{mot}$ the motor time constant, and $\boldsymbol\Omega$, $\boldsymbol\Omega_\text{ss}$ the current and commanded rotor speeds. Propulsion and aerodynamic effects are captured by $\bm{f}_\text{prop}$, $\boldsymbol{\tau}_\text{prop}$ and $\bm{f}_\text{aero}$, $\boldsymbol{\tau}_\text{aero}$. We employ a quadratic thrust model with data-driven corrections following~\cite{kaufmann_champion-level_2023}; see~\cite{bauersfeld2021neurobem} for further details.

The policy observation at time $t$ consists of position $p_t \in \mathbb{R}^3$, linear velocity $v_t \in \mathbb{R}^3$, the first two columns of the rotation matrix $R_t \in \mathbb{R}^6$, angular velocity $\omega_t \in \mathbb{R}^3$, and the previous action $a_{t-1} \in \mathbb{R}^4$. The action space consists of collective thrust and body rates (CTBR), $a_t \in \mathbb{R}^4$. The shaped reward used for the PPO baseline and synthetic preference generation is described in Appendix~\ref{app:shaped_reward}.

\subsection{Preference-based Reinforcement Learning}

To formalize preference-based reinforcement learning, we first define the concepts of \textit{Preference} and \textit{Preference Modeling}, \textit{Cross Entropy Loss}, and \textit{Synthetic Preference Generation}.\\

\textbf{Preference.} A preference label \( p \in [0,1] \) is assigned by a \textit{judge} to a pair of trajectories, reflecting the relative preference between them. We assume that preferences are determined by an underlying reward function \( r : \tau \rightarrow \mathbb{R} \), where

\begin{equation}
    p(\tau_1, \tau_2) =
\begin{cases}
0, & \text{if } r(\tau_1) > r(\tau_2), \\
0.5 & \text{if } r(\tau_1) = r(\tau_2), \\
1, & \text{if } r(\tau_1) < r(\tau_2).
\end{cases}
\end{equation}

\textbf{Preference Modeling.} A commonly used model of preference judgment, as is also being done in \cite{christiano2023deepreinforcementlearninghuman}, is viewing the reward $r$ of a trajectory $\tau$ as a latent variable that determines the choices of the annotator. More precisely, the probability of preferring one trajectory over the other is given by the softmax of the two rewards,
\begin{equation}
\hat{p}(\tau_1 > \tau_2) = \frac{\exp(\hat{r}_1)}{\exp(\hat{r}_1) + \exp(\hat{r}_2)},
\end{equation}
which is known as the Bradley–Terry (BT) model. In our notation, we refer to a predicted variable by $\hat{x}$, whereas ground truth variables are referred to as $x$.

\textbf{Cross Entropy Loss.}
In Preference PPO \cite{lee2021bprefbenchmarkingpreferencebasedreinforcement}, the reward model $\hat{r}$ is optimized by minimizing the cross entropy loss between the BT model predictions and the true labels, using the following loss:
\begin{equation}
\begin{split}
\text{loss}(\hat{r}) = - \sum_{(\tau_1^i, \tau_2^i, p^i) \in D} \Big[ 
p^i \cdot \log(\hat{p}(\tau_1^i > \tau_2^i)) \\
+ (1 - p^i) \cdot \log(\hat{p}(\tau_2^i > \tau_1^i)) \Big].
\end{split}
\label{eq:celoss}
\end{equation}

\textbf{Synthetic Preference Generation.} 
To assess the performance of our preference-based RL method, we use synthetic preferences, which serve as an oracle representing preferences based on the total handcrafted reward of each compared trajectory in the training environment~\cite{christiano2023deepreinforcementlearninghuman}. The preferred trajectory is the one that achieves a higher reward in the given task. The detailed reward functions of each environment are described in Appendix~\ref{app:shaped_reward}.

\subsubsection{Unsupervised Pretraining}\label{subsec:pretraining}
As presented in \cite{lee_pebble_2021} the pretraining step is used to improve the initial exploration of the policy. In preference-based learning, this is especially useful, as there is no guidance for the policy before the reward model has been trained. To train the reward model, some labeled trajectories are necessary. To encourage exploration, an intrinsic reward is given based on the state entropy $H(s) = \mathbb{E}_{s \sim \rho(s)} [\log(\rho(s))]$. As this is intractable, a particle-based entropy estimator \cite{liu2021behaviorvoidunsupervisedactive} is used:
\begin{equation}
    \hat{H}(s) \sim \sum_i log(\|s_i - s_i^k\|).
\end{equation}
Finally, the intrinsic reward for unsupervised exploration $r^{int}$ is given by the distance to the $k$-th nearest neighbor of the current state $s_t$:
\begin{equation}
    r^{int}(s_t) = log(\|s_t - s_t^k\|).
\end{equation}

\subsubsection{Query and Reward Training Scheduling} 
Querying (collecting new preferences) and reward-model retraining follow the same schedule: the reward model is retrained each time new preferences are gathered. Schedules are specified in terms of $n_{steps}$, the number of agent-environment interaction steps, and we define an epoch as the interval between two consecutive retraining phases. The preference-based hyperparameters are listed in Table~\ref{tab:hyperparams_pref} in Appendix~\ref{app:hyperparams}.

\subsubsection{Query Selection}
We select and pair trajectories for preference comparisons using three distinct heuristics. Each method contributes one-third of the total pairs, and each trajectory is used at most once.

\paragraph{Random} Pairs are selected at random. This serves as the default during the initial pairing phase before the reward model has been trained.

\paragraph{Ensemble Disagreement} All possible pairs are generated from the available trajectories, and each pair receives a preference label from every ensemble member. Pairs are then sorted in descending order by the number of ensemble members that disagree, and the highest-disagreement pairs are selected such that no trajectory is repeated.

\paragraph{Current with Previous Epoch} Here, an epoch refers to the interval between two reward retraining phases. A trajectory generated after the most recent reward update is paired with one generated before it.

\subsection{REC Preference PPO}
In this section, we introduce REC, a probabilistic reward learning framework that explicitly models uncertainty in the predicted reward. REC consists of three components: a probabilistic loss function for the reward model, an uncertainty-aware reward aggregation strategy, and an ensemble resetting mechanism. The full training procedure is summarized as pseudocode in Appendix~\ref{app:rec_pseudocode}.

\subsubsection{Loss Function}

\paragraph{Probabilistic Cross Entropy Loss.} For the reward model, we use an ensemble of $n_{\mathrm{ensemble}}$ (denoted $n$ for brevity) multi-layer perceptrons. Each ensemble member takes the current observation-action pair $(o_t, a_t)$ as input and outputs a scalar reward prediction $r^i_{t,\text{pred}}$. The distributional reward parameters are obtained from the ensemble statistics:
\begin{equation}
    r_{\text{mean}}(o_t, a_t) = \frac{1}{n}\sum_{i=1}^{n} r^i_{t,\text{pred}}, \quad r_{\text{std}}(o_t, a_t) = \text{std}\left(\{r^i_{t,\text{pred}}\}_{i=1}^{n}\right).
\end{equation}
We model the reward at each timestep as a sample from a normal distribution parameterized by these ensemble statistics:
\begin{equation}
    r(o_t, a_t) \sim \mathcal{N}\big(r_{\text{mean}}(o_t, a_t), r_{\text{std}}(o_t, a_t)\big).
\end{equation}
The reward for the whole trajectory is then given by
\begin{equation}
    r(\tau) \sim \sum_t\mathcal{N}\big( r_{\text{mean}}(o_t, a_t), r_{\text{std}}(o_t, a_t)\big).
\end{equation}
Assuming independence across timesteps, we can rewrite this as:
\begin{equation}
\begin{split}
    r(\tau) \sim \mathcal{N}\Bigg( \sum_t r_{\text{mean}}(o_t, a_t), \sqrt{\sum_t r_{\text{std}}(o_t, a_t)^2} \Bigg)
    \\ = \mathcal{N} \Bigg( r_{\text{mean}}(\tau), r_{\text{std}}(\tau) \Bigg).
\end{split}
\end{equation}
Given the distribution for the trajectory reward $r(\tau)$, we can model the preference between two trajectories. Here, $\Phi$ denotes the cumulative distribution function (CDF) of the standard normal, which gives the probability that a sample from $r(\tau_1)$ is larger than one from $r(\tau_2)$:
\begin{equation}
    p(\tau_1 > \tau_2) = \Phi\left( \frac{r_{\text{mean}}(\tau_1) - r_{\text{mean}}(\tau_2)}{\sqrt{r_{\text{std}}(\tau_1)^2 + r_{\text{std}}(\tau_2)^2}} \right).
\end{equation}
Note that this replaces the Bradley-Terry softmax in Eq.~\ref{eq:celoss} with a Gaussian CDF preference model that naturally incorporates reward uncertainty. When ensemble members agree and the standard deviations approach zero, this formulation recovers the deterministic preference model. We derive the preference loss by minimizing the cross-entropy with respect to this probabilistic preference:
\begin{equation}
\begin{split}
\text{loss}(\hat{r}) = - \sum_{(\tau_1^i, \tau_2^i, p^i) \in D} \Big[ 
p^i \log \hat{p}(\tau_1^i > \tau_2^i) \\
+ (1 - p^i) \log \big(1 - \hat{p}(\tau_1^i > \tau_2^i) \big) \Big].
\end{split}
\label{eq:prob_celoss}
\end{equation}
\paragraph{Standard Deviation Target Loss.} To encourage more stable standard deviation estimates across ensemble members, we introduce an additional loss term. This term penalizes the squared error between a target standard deviation $\sigma_{target}$ and the mean standard deviation across all timesteps in the preference dataset, $\bar{\sigma}$:
\begin{equation}
    \text{loss}_{\sigma} = (\bar{\sigma} - \sigma_{target})^2.
\end{equation}
Without this regularization, ensemble members can collapse to identical predictions with near-zero standard deviation, effectively recovering the deterministic Bradley-Terry model and negating the benefits of probabilistic modeling.

\vspace*{0.15cm}

\subsubsection{Reward Aggregation}

To provide the reward signal for policy optimization, we aggregate the ensemble predictions using a noisy aggregation method that increases the reward in regions of high uncertainty, following the approach of \cite{liang2022rewarduncertaintyexplorationpreferencebased}:
\begin{equation}
    r_{t,\text{pred}}^{\text{agg}} = \frac{1}{n} \sum_{i=1}^n r_{t,\text{pred}}^{i} + |X|,
\end{equation}
where $X$ is drawn from:
\begin{equation}
    X \sim \mathcal{N}\left(0, \sqrt{\frac{1}{n} \sum_{i=1}^n \left(r_{t,\text{pred}}^{i} - \frac{1}{n} \sum_{j=1}^n r_{t,\text{pred}}^{j}\right)^2}\right).
\end{equation}
The absolute value ensures that ensemble disagreement always results in a positive reward bonus, encouraging the policy to visit states where the reward model is uncertain. This drives exploration toward regions where additional preference feedback would be most informative.

\vspace*{0.15cm}

\subsubsection{Ensemble Resetting}
Before each reward model retraining, we evaluate each ensemble member separately on the newly added preferences. We then re-initialize the weights of the worst-performing $n_{reset} \in [0,\dots,n_{\mathrm{ensemble}}]$ ensemble members. This prevents poorly performing members from corrupting the ensemble statistics and helps maintain diversity among the ensemble, which is essential for meaningful uncertainty estimates. Without resetting, ensemble members can converge to similar predictions over time, reducing the effectiveness of both the probabilistic loss and the uncertainty-driven exploration. Unless otherwise specified, we use $n_{reset} = 1$ in our experiments.

%% file: chapters/results.tex
\section{Results}
\label{sec:result}

\subsection{DM Control Evaluation}
To validate REC beyond aerial robotics, we evaluate on the \textit{walker-walk} task from the DM Control Suite~\cite{tunyasuvunakool_2020}, using the implementation of Tai et al.~\cite{jun_jet_tai_2023_8140744}. We ablate each component of REC with respect to Preference PPO across 3 seeds, using the same hyperparameters as in~\cite{lee2021bprefbenchmarkingpreferencebasedreinforcement}. Fig.~\ref{fig:mujoco_ep_rew_full} shows the mean episode reward during training, and Table~\ref{tab:bpref_abl} reports the best evaluation reward per method. Each component improves over the Preference PPO baseline, with the probabilistic loss and reward noise yielding the largest gains. Ensemble resetting slightly reduces mean reward on this task but decreases variance from $\pm 55.5$ to $\pm 43.3$, indicating more consistent training. Its benefit is more pronounced on the quadrotor task (Table~\ref{tab:drone_abl}), where maintaining ensemble diversity has greater impact.

\begin{figure}[ht!]
\centering
\pgfplotstableread[col sep=comma]{figures/ral_csvs/mujoco/train_reward_bpref_gt_.csv}\mujocogtdata
\pgfplotstableread[col sep=comma]{figures/ral_csvs/mujoco/train_reward_bpref_prefppo_.csv}\mujocoprefppodata
\pgfplotstableread[col sep=comma]{figures/ral_csvs/mujoco/train_reward_bpref_prefppo_probLoss_.csv}\mujocoprefppoproblossdata
\pgfplotstableread[col sep=comma]{figures/ral_csvs/mujoco/train_reward_bpref_prefppo_probLoss_gaussExp_.csv}\mujocoprefppoproblossexpdata
\pgfplotstableread[col sep=comma]{figures/ral_csvs/mujoco/train_reward_bpref_prefppo_probLoss_gaussExp_resetEns_.csv}\mujocoprobprefppodata
\begin{tikzpicture}
\begin{axis}[
    width=\columnwidth,
    height=6cm,
    xlabel={\footnotesize{Environment Steps}},
    x tick scale label style={yshift=0.5em, xshift=1.2em},
    ylabel={\footnotesize{Evaluation Reward}},
    ylabel style={yshift=-1em},
    grid=major,
    grid style={gray!30},
    legend style={at={(0.98,0.02)}, anchor=south east, /tikz/font=\scriptsize, legend cell align=left},
    xmin=0,
    xmax=4000000,
    ymin=0,
    ymax=1000
]

\addplot [
    name path=mujoco_gt_upper,
    draw=none,
    forget plot
] table [
    x=x,
    y expr=\thisrow{mean} + \thisrow{std}
] {\mujocogtdata};

\addplot [
    name path=mujoco_gt_lower,
    draw=none,
    forget plot
] table [
    x=x,
    y expr=\thisrow{mean} - \thisrow{std}
] {\mujocogtdata};

\addplot [
    thick,
    matlab1,
    mark=none,
    forget plot
] table [
    x=x,
    y=mean
] {\mujocogtdata};

\addplot [
    fill=matlab1!40,
    fill opacity=0.5,
    draw=none
] fill between [
    of=mujoco_gt_upper and mujoco_gt_lower,
    on layer=axis background
];
\addlegendentry{PPO (shaped reward)}

\addplot [
    name path=mujoco_prefppo_upper,
    draw=none,
    forget plot
] table [
    x=x,
    y expr=\thisrow{mean} + \thisrow{std}
] {\mujocoprefppodata};

\addplot [
    name path=mujoco_prefppo_lower,
    draw=none,
    forget plot
] table [
    x=x,
    y expr=\thisrow{mean} - \thisrow{std}
] {\mujocoprefppodata};

\addplot [
    thick,
    matlab2,
    mark=none,
    forget plot
] table [
    x=x,
    y=mean
] {\mujocoprefppodata};

\addplot [
    fill=matlab2!40,
    fill opacity=0.5,
    draw=none
] fill between [
    of=mujoco_prefppo_upper and mujoco_prefppo_lower,
    on layer=axis background
];
\addlegendentry{Preference PPO}

\addplot [
    name path=mujoco_prefppo_probloss_upper,
    draw=none,
    forget plot
] table [
    x=x,
    y expr=\thisrow{mean} + \thisrow{std}
] {\mujocoprefppoproblossdata};

\addplot [
    name path=mujoco_prefppo_probloss_lower,
    draw=none,
    forget plot
] table [
    x=x,
    y expr=\thisrow{mean} - \thisrow{std}
] {\mujocoprefppoproblossdata};

\addplot [
    thick,
    matlab3,
    mark=none,
    forget plot
] table [
    x=x,
    y=mean
] {\mujocoprefppoproblossdata};

\addplot [
    fill=matlab3!40,
    fill opacity=0.5,
    draw=none
] fill between [
    of=mujoco_prefppo_probloss_upper and mujoco_prefppo_probloss_lower,
    on layer=axis background
];
\addlegendentry{Preference PPO + Prob. Loss}

\addplot [
    name path=mujoco_prefppo_probloss_exp_upper,
    draw=none,
    forget plot
] table [
    x=x,
    y expr=\thisrow{mean} + \thisrow{std}
] {\mujocoprefppoproblossexpdata};

\addplot [
    name path=mujoco_prefppo_probloss_exp_lower,
    draw=none,
    forget plot
] table [
    x=x,
    y expr=\thisrow{mean} - \thisrow{std}
] {\mujocoprefppoproblossexpdata};

\addplot [
    thick,
    matlab4,
    mark=none,
    forget plot
] table [
    x=x,
    y=mean
] {\mujocoprefppoproblossexpdata};

\addplot [
    fill=matlab4!40,
    fill opacity=0.5,
    draw=none
] fill between [
    of=mujoco_prefppo_probloss_exp_upper and mujoco_prefppo_probloss_exp_lower,
    on layer=axis background
];
\addlegendentry{Preference PPO + Prob. Loss + Gauss Exp}

\addplot [
    name path=mujoco_probprefppo_upper,
    draw=none,
    forget plot
] table [
    x=x,
    y expr=\thisrow{mean} + \thisrow{std}
] {\mujocoprobprefppodata};

\addplot [
    name path=mujoco_probprefppo_lower,
    draw=none,
    forget plot
] table [
    x=x,
    y expr=\thisrow{mean} - \thisrow{std}
] {\mujocoprobprefppodata};

\addplot [
    thick,
    matlab5,
    mark=none,
    forget plot
] table [
    x=x,
    y=mean
] {\mujocoprobprefppodata};

\addplot [
    fill=matlab5!40,
    fill opacity=0.5,
    draw=none
] fill between [
    of=mujoco_probprefppo_upper and mujoco_probprefppo_lower,
    on layer=axis background
];
\addlegendentry{REC Preference PPO (ours)}

\end{axis}
\end{tikzpicture}
\caption{Average reward on the \textit{walker-walk} task with 1000 synthetic preferences while training. Shaded areas denote standard deviation across 3 seeds. PPO with the shaped environment reward serves as the upper baseline. Ablation components are introduced incrementally to obtain REC Preference PPO.}
\label{fig:mujoco_ep_rew_full}
\end{figure}

\vspace{-0.5cm}

\begin{table}[ht]
\centering
\caption{\textnormal{Maximum evaluation reward during training on the \textit{walker-walk} task. Values represent the highest mean evaluation reward across three seeds ($\pm$ standard deviation). Components of REC are incrementally introduced starting from the Preference PPO baseline.}}
\begin{tabularx}{0.9\columnwidth}{@{}>{\raggedright\arraybackslash}X>{\raggedright\arraybackslash}X@{}}
\textbf{Method} & \textbf{Max. Eval. Rew.} \\ \hline
\grayrow
PPO & 
$930.7 \pm 12.0$ \\
Preference PPO & 
$719.0 \pm 9.3$ \\
\grayrow
\makecell[tl]{Preference PPO \\ $\quad$ + Prob. Rew. Loss} & 
$777.5 \pm 66.2$ \\
\makecell[tl]{Preference PPO \\ $\quad$ + Prob. Rew. Loss \\ $\quad$  + Rew. Noise} & 
$829.4 \pm 55.5$ \\
\grayrow
\makecell[tl]{Preference PPO \\ $\quad$ + Prob. Rew. Loss \\ $\quad$ + Rew. Noise \\ $\quad$ + Reset Ensemble --- (REC)} & 
$815.1 \pm 43.3$
\end{tabularx}
\vspace{-0.3cm}
\label{tab:bpref_abl}
\end{table}

\begin{figure*}[ht]
    \centering
    \subfloat[PPO (shaped reward)]{%
        \includegraphics[width=0.23\textwidth]{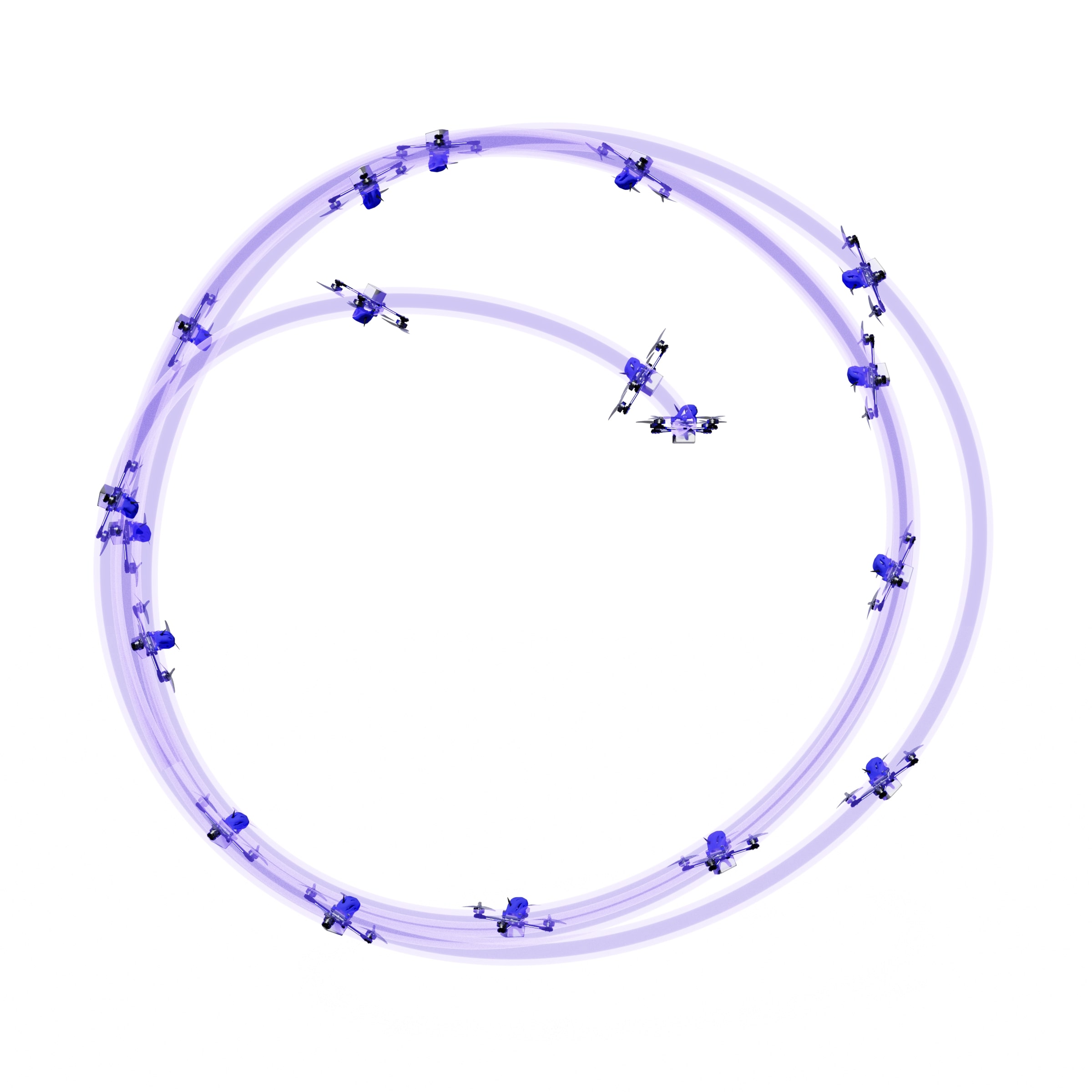}
        \label{fig:gt_stop_motion}
    }
    \hspace{0.01\textwidth}
    \subfloat[Preference PPO \\(synthetic)]{%
        \includegraphics[width=0.23\textwidth]{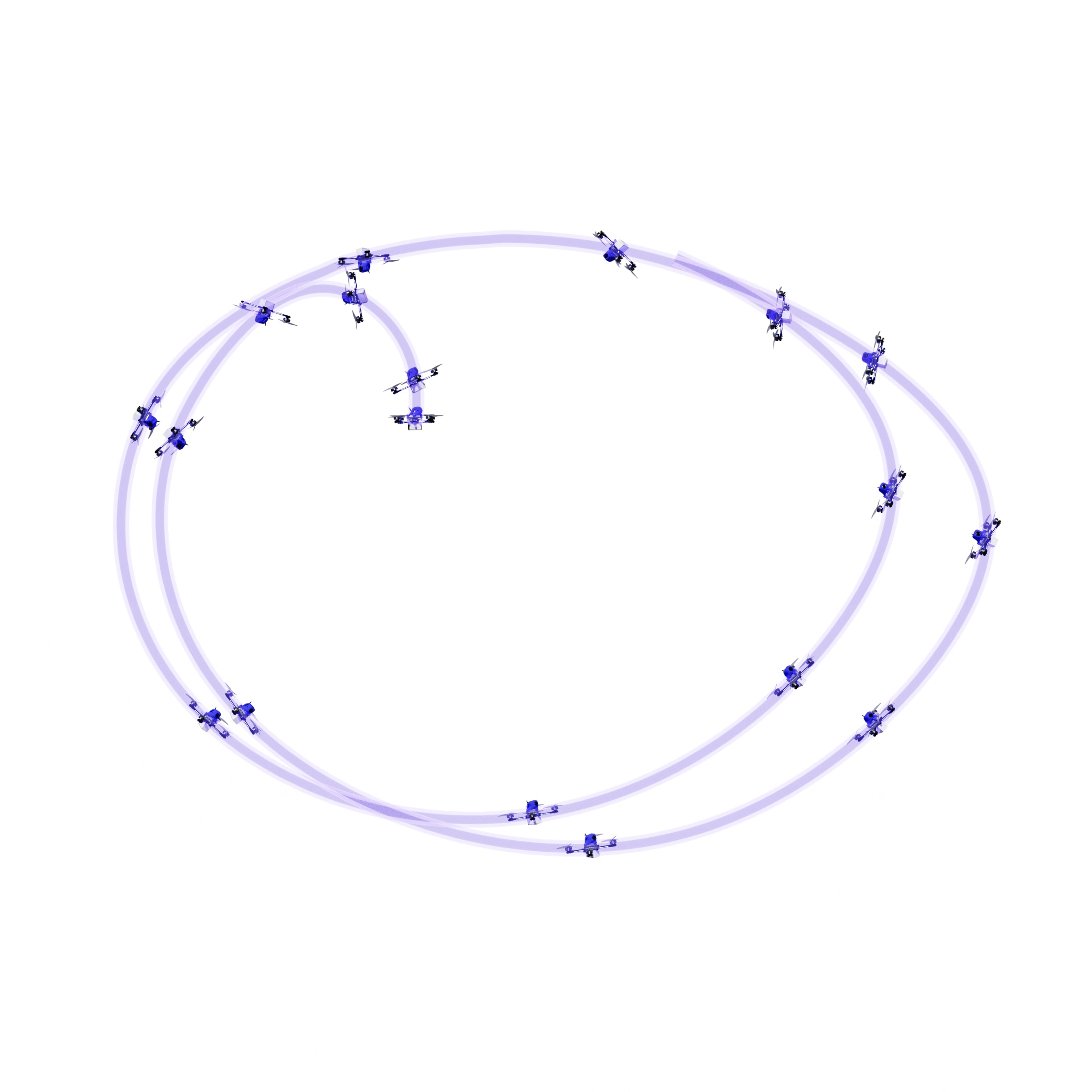}
        \label{fig:prefppo_stop_motion}
    }
    \hspace{0.01\textwidth}
    \subfloat[REC Preference PPO \\(synthetic, our algorithm)]{%
        \includegraphics[width=0.23\textwidth]{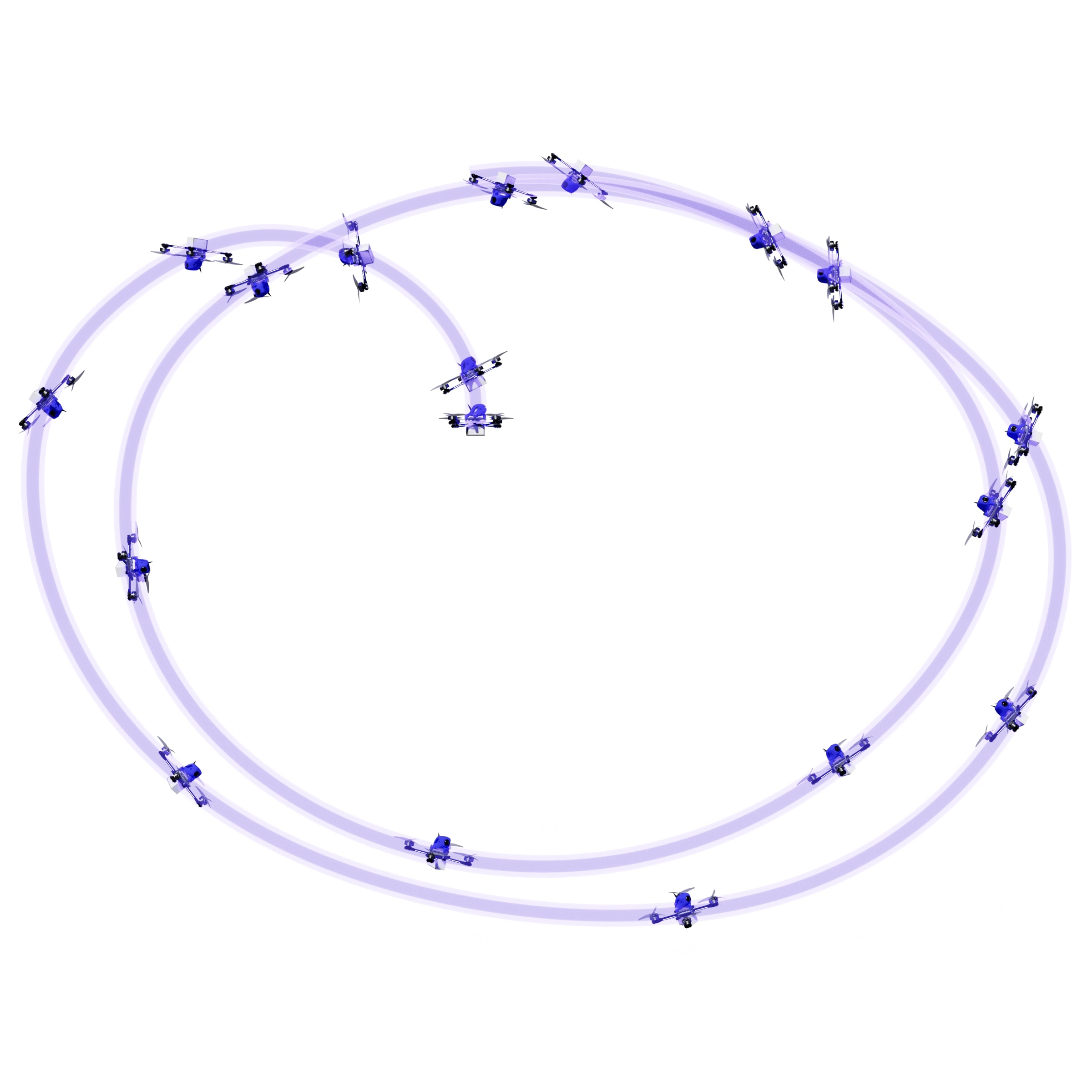}
        \label{fig:ours_stop_motion}
    }
    \hspace{0.01\textwidth}
    \subfloat[REC Preference PPO \\(human, our algorithm)]{%
        \includegraphics[width=0.23\textwidth]{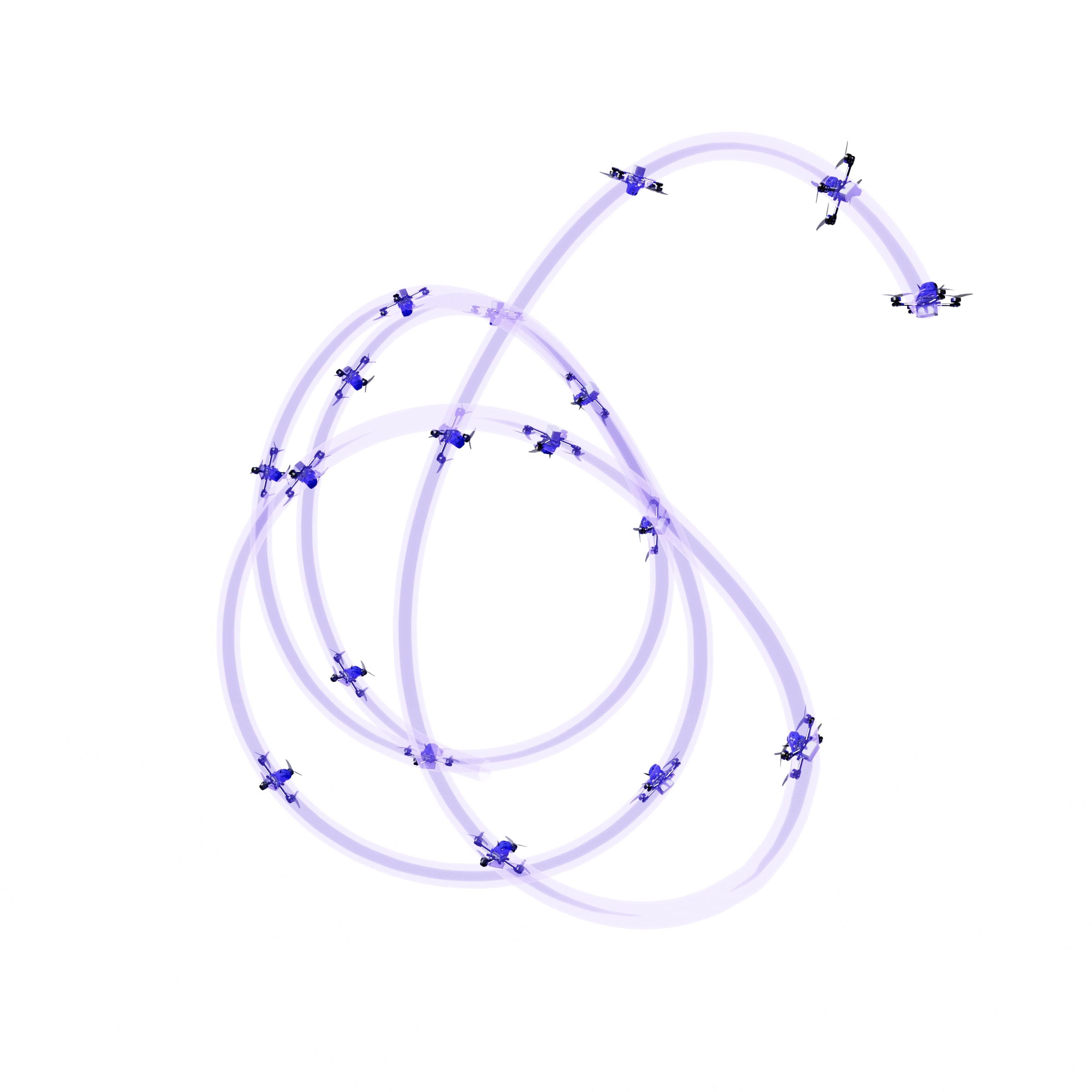}
        \label{fig:human_stop_motion}
    }

    \caption{
    Stop-motion visualizations of the highest-reward evaluation rollout in simulation for each training configuration. 
    (a)~PPO with shaped rewards. 
    (b)~Preference PPO with 1000 synthetic preferences. 
    (c)~REC Preference PPO with 1000 synthetic preferences. 
    (d)~REC Preference PPO with 1000 human-labeled preferences. 
    }
    \label{fig:stopmotion_all}
\end{figure*}

\subsection{Evaluation on the Quadrotor}
We evaluate four training configurations in the Flightmare~\cite{song2020flightmare} simulator: PPO with the shaped reward, Preference PPO, and REC Preference PPO with both synthetic and human preferences. Stop-motion visualizations of the best rollouts are shown in Fig.~\ref{fig:stopmotion_all}, and Table~\ref{tab:drone_abl} reports the mean best evaluation reward across three seeds. Fig.~\ref{fig:drone_prefppo_vs_our} shows the evaluation reward over training for the continuous powerloop task using 1000 synthetic preferences. REC Preference PPO achieves a mean evaluation reward of $382.4 \pm 80.8$, corresponding to $88.4\%$ of the shaped reward baseline ($432.4 \pm 190.8$), compared to $238.9 \pm 157.5$ ($55.2\%$) for standard Preference PPO. Notably, REC also exhibits substantially lower variance across seeds, indicating more reliable convergence on this challenging exploration problem.

\begin{figure*}[ht]
    \centering
    \subfloat[PPO (shaped reward)]{%
        \includegraphics[width=0.23\textwidth]{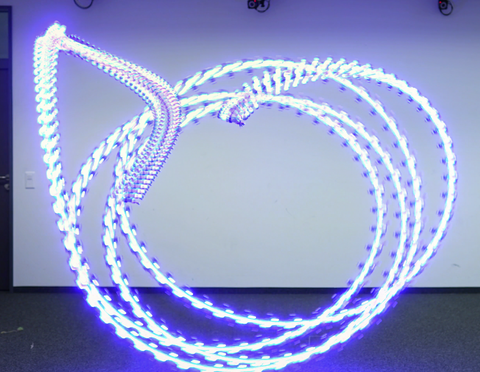}
        \label{fig:ppo_ground_truth}
    }
    \hfill
    \subfloat[Preference PPO \\(synthetic)]{%
        \includegraphics[width=0.23\textwidth]{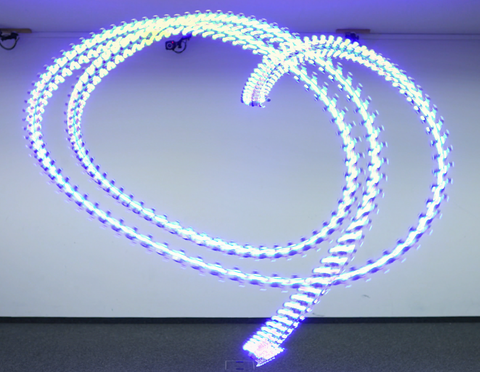}
        \label{fig:prefppo_synthetic}
    }
    \hfill
    \subfloat[REC Preference PPO \\(synthetic, our algorithm)]{%
        \includegraphics[width=0.23\textwidth]{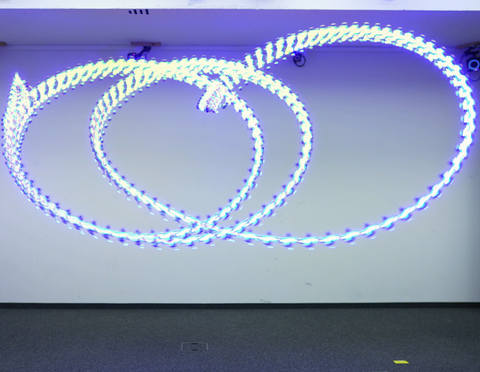}
        \label{fig:probprefppo_synthetic}
    }
    \hfill
    \subfloat[REC Preference PPO \\(human, our algorithm)]{%
        \includegraphics[width=0.23\textwidth]{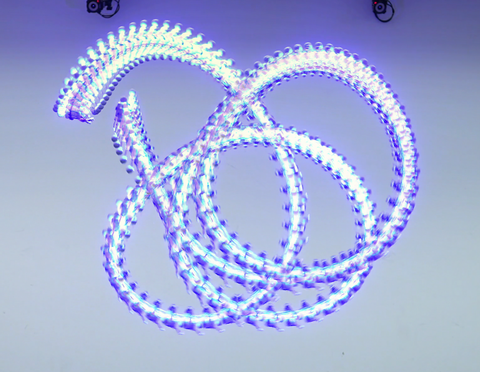}
        \label{fig:probprefppo_human}
    }
    \hspace{3em}

    \caption{Long-exposure photographs of real-world deployment across four training configurations. 
    (a)~PPO with shaped rewards. 
    (b)~Preference PPO with 1000 synthetic preferences. 
    (c)~REC Preference PPO with 1000 synthetic preferences. 
    (d)~REC Preference PPO with 1000 human-labeled preferences on the powerloop task.}
    \label{fig:deployments}
\end{figure*}

\begin{figure}[ht]
\centering
\pgfplotstableread[col sep=comma]{figures/ral_csvs/drone/drone_eval_reward_gt_baseline_.csv}\gtdata
\pgfplotstableread[col sep=comma]{figures/ral_csvs/drone/drone_eval_reward_pb_baseline_.csv}\baselinedata
\pgfplotstableread[col sep=comma]{figures/ral_csvs/drone/drone_eval_reward_pb_prefppo_.csv}\prefppodata
\begin{tikzpicture}
\begin{axis}[
    width=\columnwidth,
    height=5cm,
    xlabel={\footnotesize{Environment Steps}},
    x tick scale label style={yshift=0.5em, xshift=1.2em},
    ylabel={\footnotesize{Evaluation Reward}},
    grid=major,
    grid style={gray!30},
    legend style={
    at={(0.02,0.98)},
    anchor=north west,
    /tikz/font=\footnotesize,
    legend cell align=left,
    legend columns=1,
    inner sep=1pt,
    row sep=-2pt,
    /tikz/every even column/.append style={column sep=2pt}
    },
    xmin=0,
    xmax=30000000,
    ymin=-100,
    ymax=1000
]
\addplot [name path=gt_upper, draw=none, forget plot] table [x=x, y expr=\thisrow{mean} + \thisrow{std}] {\gtdata};
\addplot [name path=gt_lower, draw=none, forget plot] table [x=x, y expr=\thisrow{mean} - \thisrow{std}] {\gtdata};
\addplot [thick, matlab1, mark=none, forget plot] table [x=x, y=mean] {\gtdata};
\addplot [fill=matlab1!40, fill opacity=0.5, draw=none] fill between [of=gt_upper and gt_lower, on layer=axis background];
\addlegendentry{PPO (shaped reward)}
\addplot [name path=baseline_upper, draw=none, forget plot] table [x=x, y expr=\thisrow{mean} + \thisrow{std}] {\baselinedata};
\addplot [name path=baseline_lower, draw=none, forget plot] table [x=x, y expr=\thisrow{mean} - \thisrow{std}] {\baselinedata};
\addplot [thick, matlab2!80, mark=none, forget plot] table [x=x, y=mean] {\baselinedata};
\addplot [fill=matlab2!40, fill opacity=0.5, draw=none] fill between [of=baseline_upper and baseline_lower, on layer=axis background];
\addlegendentry{Preference PPO}
\addplot [name path=prefppo_upper, draw=none, forget plot] table [x=x, y expr=\thisrow{mean} + \thisrow{std}] {\prefppodata};
\addplot [name path=prefppo_lower, draw=none, forget plot] table [x=x, y expr=\thisrow{mean} - \thisrow{std}] {\prefppodata};
\addplot [thick, matlab5!, mark=none, forget plot] table [x=x, y=mean] {\prefppodata};
\addplot [fill=matlab5!40, fill opacity=0.5, draw=none] fill between [of=prefppo_upper and prefppo_lower, on layer=axis background];
\addlegendentry{REC Preference PPO (ours)}
\end{axis}
\end{tikzpicture}
\caption{Mean evaluation reward over 3 seeds during training of the continuous powerloop with 1000 synthetic preferences. Shaded areas indicate standard deviation.}
\vspace{-0.7cm}
\label{fig:drone_prefppo_vs_our}
\end{figure}

\begin{table}[ht]
\centering
\caption{\textnormal{Maximum evaluation reward during training of the continuous powerloop task. Values represent the highest mean evaluation reward across three seeds ($\pm$ standard deviation). Components of REC are incrementally introduced starting from the Preference PPO baseline.}}
\label{tab:drone_abl}
\begin{tabularx}{0.9\columnwidth}{@{}>{\raggedright\arraybackslash}X>{\raggedright\arraybackslash}X@{}}
\textbf{Method} & \textbf{Max. Eval. Rew.} \\ \hline
\grayrow
PPO (shaped reward) & 
$432.4 \pm 190.8$ \\
Preference PPO & 
$238.9 \pm 157.5$ \\
\grayrow
\makecell[tl]{Preference PPO \\ $\quad$ + Prob. Rew. Loss} & 
$281.0 \pm 187.4$ \\
\makecell[tl]{Preference PPO \\ $\quad$  + Prob. Rew. Loss \\ $\quad$  + Reward Noise} & 
$153.9 \pm 174.2$ \\
\grayrow
\makecell[tl]{Preference PPO \\ $\quad$ + Prob. Rew. Loss \\ $\quad$ + Rew. Noise \\ $\quad$ + Reset Ensemble --- (REC)} & 
$382.4 \pm 80.8$
\end{tabularx}
\vspace{-0.3cm}
\end{table}

\subsection{Preference Feedback from a Human Annotator}
We evaluate REC Preference PPO using preference labels provided by a human annotator. The experimental settings are identical to the synthetic preference experiments, with the annotator comparing trajectory pairs rendered from the simulator. The annotator has the additional option of responding \textit{`I~can't~tell'}, in which case the pair is discarded and replaced. Of the 1064 comparisons shown, 1000 valid preference labels were collected. The complete training and labeling process took 4\,h\,43\,min. Since no ground-truth reward is assumed to be available in this setting, the policy checkpoint for evaluation and deployment is selected by the annotator rather than by maximum evaluation reward. In this study, the human annotator is the same person who designed the shaped reward used for the synthetic experiments. The disagreement reported below therefore arises despite full familiarity with the intended task, which reinforces that it stems from the limitations of hand-crafted rewards rather than annotator inexperience; studying inter-annotator agreement with independent evaluators remains future work.
\begin{table}[ht]
\centering
\caption{\textnormal{Agreement between the shaped reward function and the human annotator. Rows indicate the shaped reward's preference, columns the human's choice. The shaped reward never produces ties by construction.}}
\begin{tabular}{lccc}
 & \textbf{Human: $\tau_1$} & \textbf{Human: $\tau_2$} & \textbf{Human: tie} \\ \hline
\textbf{Reward: $\tau_1$} & 323 & 167 & 37 \\
\textbf{Reward: $\tau_2$} & 196 & 238 & 39 \\
\end{tabular}
\label{fig:human_vs_gt_conf}
\vspace{-0.3cm}
\end{table}
Table~\ref{fig:human_vs_gt_conf} reports the agreement between the annotator's preferences and those implied by the shaped reward; excluding ties, the two agree only $60.7\%$ of the time. As discussed in Section~\ref{sec:result}-E, this reflects the limited ability of the shaped reward to capture what an observer values when evaluating acrobatic maneuvers rather than poor annotation quality. Despite this misalignment, the policy trained from human preferences executes the powerloop both in simulation and on the real platform (Section~\ref{sec:deployment}), indicating that preference feedback captures task-relevant information that the shaped reward does not fully encode.

\begin{figure}[ht]
    \centering
    \includegraphics[width=0.8\linewidth]{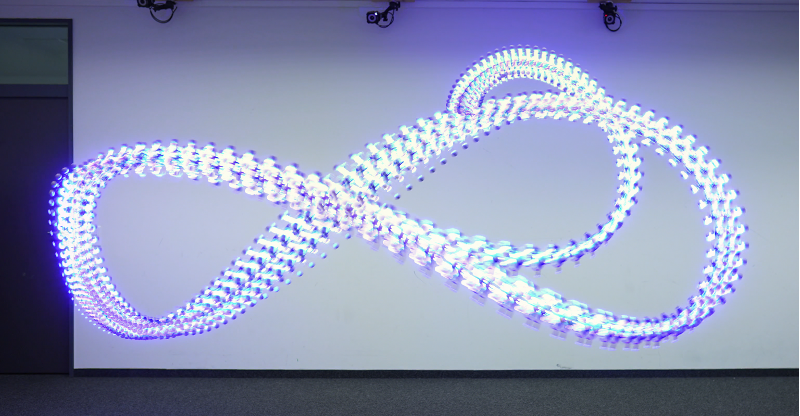}
    \caption{Vertical Figure-8 (double powerloop) trained purely from human preference feedback using REC Preference PPO without any handcrafted reward function.}
    \vspace{-0.7cm}
    \label{fig:novelskill}
\end{figure}
\vspace*{-0.2cm}
\subsection{Real-World Deployment and Novel Acrobatic Skills}\label{sec:deployment}
We deploy the simulation-trained policies on a 220\,g
quadrotor with a static thrust-to-weight ratio of 6.5, commanding collective thrust and body rates (CTBR) at 50\,Hz. For configurations trained with synthetic preferences, we select the checkpoint with the highest evaluation reward. For the human preference experiment, the annotator selects the checkpoint. All policies are transferred zero-shot without any real-world fine-tuning. Deployment results for each training configuration are shown in Fig.~\ref{fig:deployments}. All four configurations successfully execute continuous powerloop maneuvers on the real platform, 
demonstrating that both synthetic and human preferences enable sim-to-real transfer of agile flight. To show the framework's generality, we further train a novel vertical Figure-8 (double powerloop) from human preferences alone, without changing any hyperparameters. The resulting policy, shown in Fig.~\ref{fig:novelskill}, executes this maneuver on the real drone, illustrating how preference feedback lets non-expert users specify new skills without reward engineering.

\subsection{Discussion}

The ablations on both walker-walk and the quadrotor indicate that the probabilistic loss and reward noise contribute the largest individual gains, while ensemble resetting primarily improves consistency; its benefit scales with task difficulty, having a modest effect on walker-walk but a larger one on the quadrotor, where agents frequently fail to explore beyond an initial half-flip. The $60.7\%$ agreement between the shaped reward and human preferences does not indicate poor annotation but rather a fundamental limitation of manual reward engineering: observers judge maneuvers holistically through qualities such as smoothness, timing, and visual impression that are difficult to encode as Markovian terms, yet the human-trained policy still transfers successfully to the real platform. Human evaluation also has limits of its own: annotators perceive geometric trajectory properties but not control-level quantities such as body rates or energy, and the assessment is viewpoint-dependent, so multi-perspective evaluation or control-aware visual feedback is a promising direction. Finally, the high seed variance reflects the exploration difficulty of acrobatic tasks in the absence of a shaping reward, which curriculum or more targeted exploration strategies could mitigate.

%% file: chapters/conclusion.tex
\section{Conclusion}
\label{sec:conclusion}

In this work, we propose REC, a probabilistic reward learning framework for preference-based reinforcement learning that explicitly models reward uncertainty through ensemble distributional predictions. Applied to acrobatic quadrotor control, REC achieves $88.4\%$ of shaped reward performance compared to $55.2\%$ for standard Preference PPO, while reducing training variance across seeds. Policies trained with both synthetic and human preference feedback transfer zero-shot to a real 220\,g quadrotor, executing continuous powerloops and a novel vertical Figure-8 maneuver without any manually designed reward function or changes in hyperparameters. The observed $60.7\%$ agreement between shaped rewards and human preferences further indicates that manual reward engineering may be insufficient for tasks where desired behavior is more naturally expressed through comparative judgments.

Several directions remain for future work. The preference collection process would benefit from multi-annotator studies and reduced viewpoint dependency to better characterize the consistency and robustness of preference signals. More broadly, extending REC to a wider range of agile flight maneuvers and robotic domains beyond quadrotors would further establish the generality of probabilistic preference-based learning.

\vspace{-0.5cm}

%% file: chapters/appendix.tex
\vspace{-0.2cm}

\section*{APPENDIX}
\setcounter{table}{0}
\renewcommand{\thesubsection}{\Alph{subsection}}
\renewcommand{\thetable}{\Alph{subsection}\arabic{table}}

\subsection{Quadrotor Powerloop Shaped Reward}
\label{app:shaped_reward}
We design a shaped, Markovian reward for the PPO baseline and for synthetic preference generation. The reward consists of three components.

\paragraph{Planar Penalty.} Encourages movement in the $xz$-plane using the L1 distance to the plane. The drone's position is given by $\vec{p}_{drone}$, with $p_{planar}$ denoting the distance from the drone to the circle plane defined by center $\vec{c}_{circle}$ and normal $\vec{n}_{plane}$:
\begin{equation}
    p_{planar} = | (\vec{p}_{drone} - \vec{c}_{circle}) \cdot \vec{n}_{plane} |_{1}.
\end{equation}

\paragraph{Circular Motion Reward.} Rewards tangential movement along a target circle. The vector $\vec{t}$ is the tangent of the circle and $\vec{v}_{circular,sat}$ is the drone's velocity projected onto the circle plane and saturated at a maximum value:
\begin{equation}
    r_{circ} = \vec{t} \cdot \vec{v}_{circular, sat}.
\end{equation}

\paragraph{Action Regularization.} Penalizes the high-frequency commands $c_{i,\text{high}} = c_i - \mathrm{LP}(c_i)$, the raw minus the low-pass filtered command, for each action dimension $i$:
\begin{equation}
    p_{reg,i} = |c_{i,\text{high}}| + |c_{i,\text{high}}|^2.
\end{equation}
The total reward is a weighted sum of these components:
\begin{equation}
    r_{total} = \alpha \cdot r_{circ} - \beta \cdot p_{planar} - \sum_i \gamma_i \cdot p_{reg,i},
\end{equation}
where $\alpha = 0.5$, $\beta = 3$, $\gamma_{\omega_{xy}} = \gamma_{\omega_{z}} = 0.008$, $\gamma_{c} = 0.0001$.

\subsection{Hyperparameters}
\label{app:hyperparams}
The hyperparameters used in our experiments are summarized in Tables~\ref{tab:hyperparams_pref} and~\ref{tab:hyperparams_ppo}, listing the preference-based and PPO settings, respectively.

\renewcommand{\arraystretch}{1.1}
\begin{table}[H]
    \centering
    \footnotesize
    \caption{\textnormal{Preference-based hyperparameters in the \textit{walker-walk} and \textit{flightmare} environments. Pairing is 50\% ensemble disagreement and 50\% random.}}
    \label{tab:hyperparams_pref}
    \begin{tabularx}{0.9\columnwidth}{@{}>{\raggedright\arraybackslash}X>{\raggedright\arraybackslash}X>{\raggedright\arraybackslash}X@{}}
        \toprule
        \textbf{Hyperparameter} & \textbf{flightmare} & \textbf{walker-walk} \\
        \midrule
        $n_{\mathrm{prefs}}$ & 1000 & 1000 \\
        $T_{\mathrm{clip}}$ & 1.25\,s & 1.25\,s \\
        $f_{\mathrm{init}}$ & 0.2 & 0.1 \\
        $n_{\mathrm{query}}$ & 10 & 10\\
        $\Delta t_{\mathrm{retrain}}$ & 400K & 16K \\
        $n_{\mathrm{ensemble}}$ & 5 & 3 \\
        $n_{\mathrm{reset}}$ & 1 & 1 \\
        $\sigma_{\mathrm{target}}$ & 0.05 & 0.05 \\
        $d_{\mathrm{hidden}}^{\mathcal{R}}$ & 256 & 256\\
        $n_{\mathrm{layers}}^{\mathcal{R}}$ & 2 & 2\\
        $\phi^{\mathcal{R}}$ & Tanh & Tanh\\
        $n_{\mathrm{epochs}}^{\mathcal{R}}$ & 100 & 100\\
        $\eta^{\mathcal{R}}$ & 0.0003 & 0.0003\\
        \bottomrule
    \end{tabularx}
    \vspace{-0.2cm}
\end{table}
\renewcommand{\arraystretch}{1.0}

\begin{table}[H]
    \centering
    \footnotesize
    \caption{\textnormal{PPO hyperparameters in the \textit{walker-walk} and \textit{flightmare} environments.}}
    \label{tab:hyperparams_ppo}
    \begin{tabularx}{0.9\columnwidth}{@{}>{\raggedright\arraybackslash}X>{\raggedright\arraybackslash}X>{\raggedright\arraybackslash}X@{}}
        \toprule
        \textbf{Hyperparameter} & \textbf{flightmare} & \textbf{walker-walk} \\
        \midrule
        $n_{\mathrm{envs}}$ & 50 & 32 \\
        $n_{\mathrm{timesteps}}$ & 30M & 4M \\
        $\gamma$ & 0.995 & 0.99 \\
        $\lambda_{\mathrm{GAE}}$ & 0.95 & 0.92 \\
        $n_{\mathrm{epochs}}$ & 10 & 20 \\
        $n_{\mathrm{steps}}$ & 250 & 500 \\
        $\epsilon_{\mathrm{clip}}$ & 0.4 & 0.4 \\
        $n_{\mathrm{batch}}$ & 12{,}500 & 64 \\
        $\log \sigma_{\mathrm{init}}$ & $-0.8$ & $0.0$ \\
        $\pi_{\mathrm{arch}}$ & [128, 128] & [256, 256, 256] \\
        $\pi_{\mathrm{activation}}$ & Tanh & Tanh \\
        $c_{\mathrm{entropy}}$ & 0.001 & 0.0 \\
        \bottomrule
    \end{tabularx}
        \vspace{-0.2cm}
\end{table}

\input{chapters/pseudocode}

%% file: chapters/pseudocode.tex
\subsection{Pseudocode}
\label{app:rec_pseudocode}
The full training procedure is summarized below.
\begin{figure}[!htb]
\centering
\footnotesize
\setlength{\tabcolsep}{0.3em}
\renewcommand{\arraystretch}{1.1}
\begin{tabularx}{\columnwidth}{@{}rX@{}}
\toprule
\multicolumn{2}{@{}l@{}}{\textbf{Algorithm 1: REC Pseudocode}}\\
\midrule
\multicolumn{2}{@{}p{0.95\columnwidth}@{}}{\textbf{Input:} env.\ $\mathcal{E}$, policy $\pi_\theta$, reward ensemble $\{\hat r_{\phi_i}\}_{i=1}^{n}$, preference budget $B$, clip length $H$, annotator $\mathcal{A}$, target std.\ $\sigma_{\mathrm{target}}$}\\
\multicolumn{2}{@{}p{0.95\columnwidth}@{}}{\textbf{Initialize:} trajectory memory $\mathcal{M}\!\leftarrow\!\emptyset$, preference dataset $\mathcal{D}\!\leftarrow\!\emptyset$}\\
\midrule
1 & Collect diverse initial trajectories by unsupervised exploration; \newline store clips of length $H$ in $\mathcal{M}$.\\
2 & \textbf{while} training budget not exhausted \textbf{do}\\
3 & \quad Select informative pairs $(\tau_0,\tau_1)$ from $\mathcal{M}$ using the reward ensemble.\\
4 & \quad Query $\mathcal{A}$ for labels $p\in\{0,0.5,1\}$; add $(\tau_0,\tau_1,p)$ to $\mathcal{D}$.\\
5 & \quad Reinitialize worst-performing ensemble member on new labels.\\
6 & \quad \textbf{for} each minibatch $(\tau_0,\tau_1,p)\sim\mathcal{D}$ \textbf{do}\\
7 & \quad\quad Predict per-timestep rewards $\hat r_{\phi_i}(o_t,a_t)$, $i=1,\dots,n$.\\
8 & \quad\quad Trajectory statistics: \newline \centerline{$\mu(\tau)=\sum_t \tfrac{1}{n}\sum_i \hat r_{\phi_i}, \quad \sigma(\tau)^2=\sum_t \operatorname{std}_i(\hat r_{\phi_i})^2$.}\\
9 & \quad\quad REC preference: \newline \centerline{$\hat p(\tau_0\!>\!\tau_1)=\Phi\!\big(\frac{\mu(\tau_0)-\mu(\tau_1)}{\sqrt{\sigma(\tau_0)^2+\sigma(\tau_1)^2}}\big)$.}\\
10 & \quad\quad Update $\{\phi_i\}$ via cross-entropy$(\hat p,p)+(\bar{\sigma}-\sigma_{\mathrm{target}})^2$.\\
11 & \quad \textbf{end for}\\
12 & \quad Collect rollouts with $\hat r^{\mathrm{agg}}_t=\tfrac{1}{n}\sum_i\hat r_{\phi_i}+|X_t|$, $\ X_t\!\sim\!\mathcal{N}(0,\operatorname{std}_i(\hat r_{\phi_i}))$.\\
13 & \quad Update $\pi_\theta$ with PPO; append new clips to $\mathcal{M}$.\\
14 & \textbf{end while}\\
\midrule
\multicolumn{2}{@{}p{0.95\columnwidth}@{}}{\textbf{Output:} trained policy $\pi_\theta$ and reward ensemble $\{\hat r_{\phi_i}\}_{i=1}^{n}$}\\
\bottomrule
\end{tabularx}
\end{figure}